\documentclass[sigconf]{acmart}
\usepackage{balance}
\usepackage{graphicx}
\usepackage{array}
\usepackage{amsfonts}
\usepackage{bm}
\usepackage{color}

\definecolor{darkolivegreen}{rgb}{0.33, 0.42, 0.18}
\definecolor{darkpastelgreen}{rgb}{0.01, 0.75, 0.24}
\definecolor{mediumjunglegreen}{rgb}{0.11, 0.21, 0.18}

\newtheorem{thm}{Theorem}

\AtBeginDocument{%
  \providecommand\BibTeX{{%
    \normalfont B\kern-0.5em{\scshape i\kern-0.25em b}\kern-0.8em\TeX}}}

\setcopyright{acmcopyright}

\copyrightyear{2022} 
\acmYear{2022} 
\setcopyright{acmcopyright}\acmConference[MM '22]{Proceedings of the 30th ACM International Conference on Multimedia}{October 10--14, 2022}{Lisboa, Portugal}
\acmBooktitle{Proceedings of the 30th ACM International Conference on Multimedia (MM '22), October 10--14, 2022, Lisboa, Portugal}
\acmPrice{15.00}
\acmDOI{10.1145/3503161.3547866}
\acmISBN{978-1-4503-9203-7/22/10}

\begin{document}\sloppy

\title{Semantic Data Augmentation based Distance Metric Learning for Domain Generalization}

\author{Mengzhu Wang$^*$}
\affiliation{%
	\institution{Alibaba Group}
	\country{Beijing, China}}
\email{wangmengzhu.wmz@alibaba-inc.com}

\author{Jianlong Yuan$^*$}
\affiliation{%
	\institution{Alibaba Group}
	\country{Beijing, China}}
\email{gongyuan.yjl@alibaba-inc.com}

\author{Qi Qian}
\affiliation{%
	\institution{Alibaba Group}
	\country{Beijing, China}}
\email{qi.qian@alibaba-inc.com}

\author{Zhibin Wang$^{\dagger}$}
\affiliation{%
	\institution{Alibaba Group}
	\country{Beijing, China}}
\email{zhibin.waz@alibaba-inc.com}

\author{Hao Li}
\affiliation{%
	\institution{Alibaba Group}
	\country{Beijing, China}}
\email{lihao.lh@alibaba-inc.com}

\makeatletter
\def\authornotetext#1{
	\if@ACM@anonymous\else
	\g@addto@macro\@authornotes{
		\stepcounter{footnote}\footnotetext{#1}}
	\fi}
\makeatother
\authornotetext{Equal Contribution.}
\authornotetext{Corresponding author.}

\renewcommand{\shortauthors}{Mengzhu Wang et al.}

\begin{abstract}
Domain generalization (DG) aims to learn a model on one or more different but related source domains that could be generalized into an unseen target domain. Existing DG methods try to prompt the diversity of source domains for the model's generalization ability, while they may have to introduce auxiliary networks or striking computational costs. On the contrary, this work applies the implicit semantic augmentation in feature space to capture the diversity of source domains. Concretely, an additional loss function of distance metric learning (DML) is included to optimize the local geometry of data distribution. Besides, the logits from cross entropy loss with infinite augmentations is adopted as input features for the DML loss in lieu of the deep features. We also provide a theoretical analysis to show that the logits can approximate the distances defined on original features well. Further, we provide an in-depth analysis of the mechanism and rational behind our approach, which gives us a better understanding of why leverage logits in lieu of features can help domain generalization. The proposed DML loss with the implicit augmentation is incorporated into a recent DG method, that is, Fourier Augmented Co-Teacher framework (FACT). Meanwhile, our method also can be easily plugged into various DG methods. Extensive experiments on three benchmarks (Digits-DG, PACS and Office-Home) have demonstrated that the proposed method is able to achieve the state-of-the-art performance. 

\end{abstract}


\maketitle

\section{Introduction}

With the rapid development of deep neural networks~\cite{sze2017efficient,montavon2018methods,yang2021deconfounded}, the performance of many essential tasks in multimedia~\cite{yang2018person,yang2020learning,yang2020tree,yang2020weakly} has been dramatically improved. However, the domain shift problem~\cite{weiss2016survey,wang2021confidence}, where the training (source domain) and test (target domain) datasets follow different distributions, is still challenging for deep learning. It is due to the fact that deep learning is easy to fit the training set with an over-parameterization neural network. The overfitting phenomenon makes the generalization to the test set from a different distribution intractable. For example, a model trained on a dataset collected from clear weather may perform poorly and unexpectedly on another dataset collected from foggy or snowy weather.

\begin{figure}[tb]
	\centering
	\includegraphics[width=\linewidth]{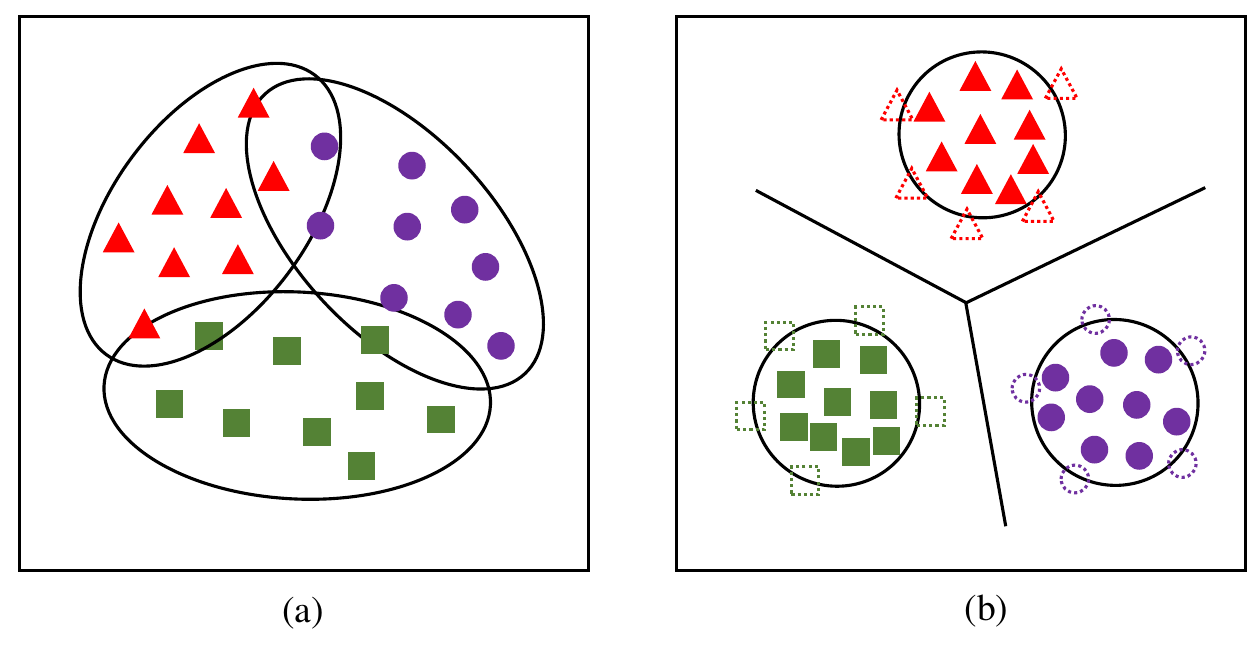}
	\caption{(a) Data distribution by CE loss; (b) Our method, where the dashed samples represent our augmented samples.}
	\label{fig1}
\end{figure}

To address the domain shift problem, domain adaptation~\cite{sun2016return,pan2010domain,cui2020towards,wang2022attention} commits to bridging the difference between the source domain and target domain with the assistance of labeled source samples but unlabeled target ones. However, in many practical applications, the unlabeled target samples are unavailable or unseen. A more realistic research topic, domain generalization (DG), is therefore intensively studied~\cite{li2018deep,li2018domain,zhou2020learning,wang2021interbn}. DG aims to leverage one or more different but related source domains to obtain an applicable model that could be generalized into an unseen target domain. 

Recent studies on DG~\cite{nuriel2020permuted,qiao2020learning,xu2020robust} indicate that the performance mainly depends on the number of diverse source samples. To this end, existing DG approaches usually utilize generative adversarial networks (GANs)~\cite{zhou2020learning} or adaptive instance normalization (AdaIN)~\cite{zhou2020learning} to synthesize unseen samples and improve the diversity of source domains accordingly. However, GANs-based DG models are difficult to be optimized. Besides, the discrimination ability of augmented samples generated by AdaIN-based DG methods is poor for instance normalization procedure throws away discriminative information.

Besides, many DG methods have variants of cross entropy (CE) loss for optimization, which is hard to capture the intra-class variance~\cite{QianSSHTLJ19}. Therefore, some method proposes to include an additional loss from distance metric learning (DML) to learn fine-grained patterns within each class~\cite{dou2019domain}. Compared with the CE loss, that of DML focuses on optimizing the local geometry of data distribution, which can obtain informative representations to model the variance in each class. The DML loss can improve the performance of DG substantially~\cite{dou2019domain} while the augmentation strategy has less been investigated for DML in DG. 

Inspired by recent study on data augmentation~\cite{wang2021regularizing}, this paper utilizes the semantic direction based data augmentation in feature space for DG to improve the diversity of source domains. Specifically, there are many semantic directions in the deep feature space based on the intriguing observation that the features are usually linearized~\cite{wang2021regularizing,li2021semantic}. Therefore, we translate a sample in source domain (s) along these semantic directions to produce more augmented samples with the same class label but different semantics as in ISDA~\cite{wang2021regularizing}. For example, we use the semantic direction corresponding to the semantic translation of "lipstick", to realize data augmentation for the feature of a person without "lipstick", then the produced samples belong to the same person but with lipstick. Compared with the conventional augmentation in input space, the semantic augmentation in feature space is more effective to generate diverse examples with the large variance for DG.

However, feature augmentation strategy of ISDA~\cite{wang2021regularizing} is only for CE loss on classification. It perturbs the logits defined by deep features and a decoupled fully-connected (FC) layer with the semantic directions from the corresponding covariance matrix. Considering that the distance in DML is defined on pairs of original examples, the coupled variables make applying the semantic augmentation for a DML loss in DG challenging. By further investigating the structure of CE loss, our analysis shows that the distance between pairs of examples with deep features can be approximated well by that with logits from CE loss. Therefore, we propose to apply the logits with semantic augmentation as input features for an arbitrary DML loss in DG. This strategy is simple yet effective to augment deep features sufficiently with semantic direction for DML and improve the generalization of DG accordingly. Fig.~\ref{fig1} (b) illustrates the proposed method. With the implicit augmentation for DML, the learned model can capture the intra-class variance better.

To verify the effectiveness of the proposed method, we apply a recent work Fourier Augmented Co-Teacher (FACT)~\cite{xu2021fourier} as a baseline DG method and include an additional DML loss with the proposed semantic augmentation strategy for it. Our experimental results have demonstrated that DML with semantic augmentation can effectively improve the domain generalization ability. In a nutshell, our main contributions are summarized as follows:
\begin{itemize}
    \item A novel implicit semantic augmentation strategy is proposed for DML in DG. Concretely, the logits with semantic augmentation is applied as input features for the DML loss in lieu of the original deep features. The theoretical analysis shows that the logits can approximate the distances defined on original features well but is much easier to be augmented by infinite augmentations in feature space.
    \item It is noteworthy that our method relies entirely on the backbone network for its implementation and computation, which makes it easy to implement, as well as computationally efficient.
    \item Extensive empirical evaluations on several competitive benchmarks including Digits-DG, PACS and Office-Home demonstrate that our method improves the baseline model by significant margins.  
\end{itemize}

\section{RELATED WORK}
\subsection{Domain Adaptation}

\begin{figure*}[t]
	\centering
	\includegraphics[width=\linewidth]{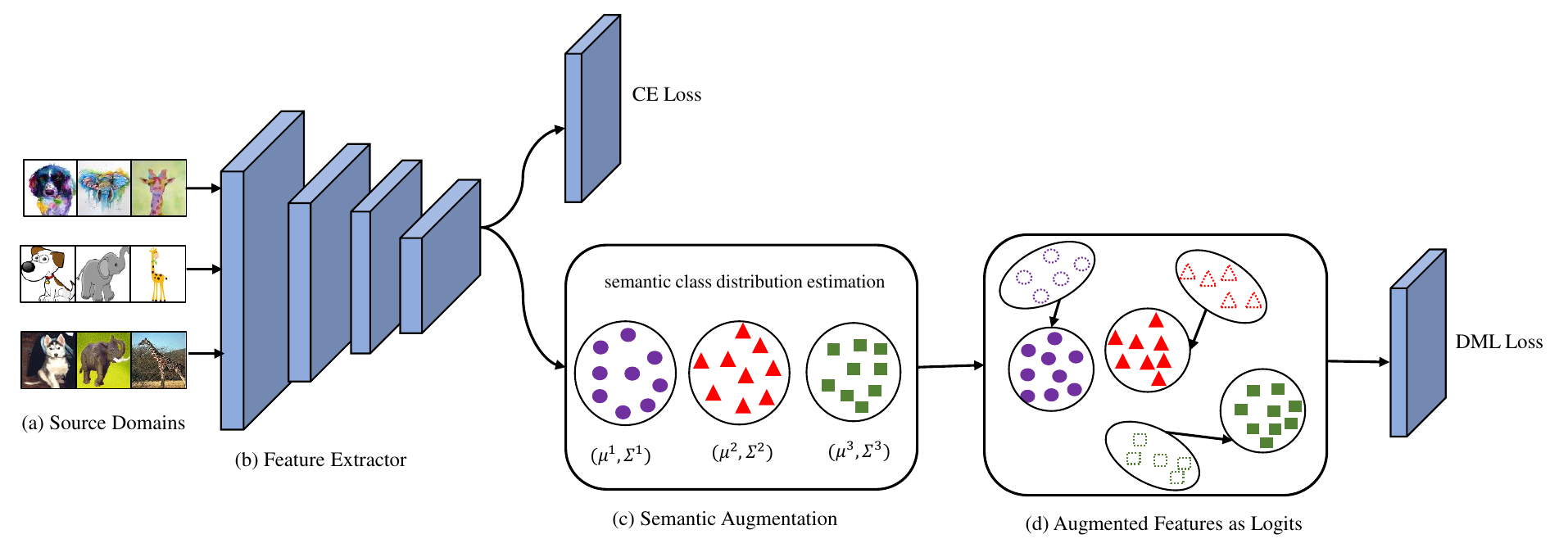}
	\caption{We augment the source features based on the mean and covariance of the intra-class features. We can also replace the original features with the logits from the semantic augmentation instead of explicitly generating augmented features. Our methods can achieve meaningful semantic transformation as shown by the augmented source features with different shapes, colors, and backgrounds.} 
	\label{fig:framework}
\end{figure*}

Semi-supervised domain adaptation methods work by imposing constraints on unlabeled or lightly labeled target domains. For instance, Ao \textit{et al}.~\cite{ao2017fast} align source and target domains by generating pseudo-labels on the unlabeled data of the target domain. Donahue \textit{et al}.~\cite{donahue2013semi} transferred knowledge by establishing similarity graph constraints on partial data of unlabeled target domain and using a projected model transfer method. Satio \textit{et al}.~\cite{saito2019semi} estimated class-specific prototypes with sparsely labeled examples of the target domain through an adversarial learning method. Yao \textit{et al}.~\cite{yao2015semi} combine subspace learning and semi-supervised learning into the domain adaptation problem to learn a subspace so that the data distribution of different domains is basically matched after mapping to this subspace.

In unsupervised domain adaptation, most methods perform feature alignment between source and target domains. To this end, PFAN~\cite{chen2019progressive} aligns discriminative features across domains by exploiting the intra-class discriminative feature of the target domain efficiently. CORAL~\cite{sun2016return} minimizes the distance between the two domains by the covariance matrices. CyCADA~\cite{hoffman2018cycada} adapts the domains by specifying the transfer between domains through a specific discriminative training task and avoiding divergence by enforcing coherence of relevant semantics before and after adaptation, resulting in spatial alignment of generated images and spatial alignment of latent representations. Recently, data augmentation~\cite{li2021transferable,kim2020learning} is another generative stream that can enhance the diversity of the source domain. Inspired by data augmentation, 
TSA~\cite{li2021transferable} augment source features with random directions sampled from the distribution class-wisely. LTIR~\cite{kim2020learning} uses a style transfer algorithm to diversity the textures of the synthesized images. 
Our method is related to domain randomization which tries to generate diverse source domains. However, they are not suitable for domain generalization as the target domain is not available. On contrary, our method is able to leverage domain randomization to solve the problem of DML and improve the diversity of the features. 

\subsection{Domain Generalization}
Domain Generalization (DG)~\cite{zhou2020learning,muandet2013domain,QianZTJSL19,ghifary2016scatter,li2018deep,shao2019multi} aims to extract knowledge from multiple source domains so that it can generalize well to unseen target domains. Unlike UDA, target domain data is not accessible during training, which makes the task more challenging than UDA. Moreover, the conditional distributions of multiple source domains are not the same.  Early DG research mainly exploits the idea of distribution alignment in domain adaptation to learn domain-invariant features through kernel methods~\cite{muandet2013domain,ghifary2016scatter} or domain adversarial learning~\cite{li2018deep,li2018domain,shao2019multi}. Later on, most DG methods try to learn the domain invariant representation and leverage data augmentation to expand the diversity of the source domain. The key idea behind the former category is to learn a domain-invariant feature by reducing the difference between representations from multiple source domains.
Different from these methods, we strive to explore the class distribution as a composite representation to expand the diversity of the class. We propose to leverage the implicit semantic augmentation strategy for DML in DG.

\section{The Proposed Method}
\subsection{Domain Generalization with Distance Metric Learning}
For the domain generalization (DG) task, there are multiple datasets of $K$ source domains $\mathcal{D}=\left\{\mathcal{D}_{1}, \mathcal{D}_{2}, \ldots, \mathcal{D}_{K}\right\}$ for training. Each dataset $\mathcal{D}_{i}$ contains a set of images $X^{i}=\left\{x_{1}^{i}, x_{2}^{i}, \ldots, x_{n_{i}}^{i}\right\}$ with the corresponding set of class labels $Y^{i}=\left\{y_{1}^{i}, y_{2}^{i}, \ldots, y_{n_{i}}^{i}\right\}$, where $n_{i}$ is the number of images in the $i^{th}$ dataset, and $y_{n_{i}}^{i} \in\{1, \ldots, C\}$, where all the datasets share the same label space. DG aims to train a network that can generalize to the unseen target domain well. Many sophisticated methods~\cite{xu2021fourier} have been developed for DG task and the optimization problem can be written as 
\begin{equation}
\label{eq:dg}
\begin{aligned}
\min _{\theta} \sum_{i} \mathcal{L}_{D G}\left(X^{i}, Y^{i} ; \theta\right)
\end{aligned}
\end{equation}
Cross entropy (CE) loss is widely applied in $\mathcal{L}_{DG}$ to obtain discriminative models as
\begin{equation}
\label{eq:celoss}
\begin{aligned}
\mathcal{L}_{CE}(\boldsymbol{W},\boldsymbol{b}) = -\log\frac{e^{\boldsymbol{w}_{y_{i}}^{\mathrm{T}} {\boldsymbol{f}}_{i} + b_{y_{i}}}}{\sum_{j=1}^{C} e^{\boldsymbol{w}_{j}^{\mathrm{T}} {\boldsymbol{f}}_{i} + b_{j}}}
\end{aligned}
\end{equation}
where $\boldsymbol{b}=\left[b_{1}, \ldots, b_{C}\right]^{\mathrm{T}} \in \mathcal{R}^{{C}}$ and $\boldsymbol{W}=\left[\boldsymbol{w}_{1}, \ldots, \boldsymbol{w}_{C}\right]^{\mathrm{T}} \in \mathcal{R}^{C \times d}$ are the biases and weight matrix corresponding to the final connected layer, respectively.

However, cross entropy loss is hard to capture the intra-class variance~\cite{QianSSHTLJ19}, which is important for generalizing learned models to the unseen domain. To preserve the diversity within each class, a loss function for distance metric learning (DML) can be included for DG~\cite{dou2019domain} and the problem can be cast as
\begin{equation}
\label{eq:overall}
\begin{aligned}
\min_{\theta} \sum_{i}\mathcal{L}_{DG}(X^i, Y^i; \theta) + \alpha \mathcal{L}_{DML}(X^i, Y^i;\theta)
\end{aligned}
\end{equation}

Unlike cross entropy loss which pulls all examples from the same class together, DML optimizes the distribution of nearest neighbors for an anchor example, which is flexible to model the intra-class variance.

Despite the success of DML in DG, augmentation, especially the augmentation in semantic space, has been less investigated for DML. Different from cross entropy loss, many DML losses are defined on triples consisting of original examples, which makes the semantic augmentation challenging. Inspired by the augmentation for classification, we develop a novel strategy for implicit data augmentation in DML to improve the performance of DG.

\subsection{Semantic Data Augmentation for Classification} 
First, we briefly review the semantic augmentation for classification~\cite{wang2021regularizing}. Compared with augmentation in the original input space, that in feature space can be more effective. The main challenge is to obtain appropriate translation directions in the feature space. To alleviate the challenge, In the proposed algorithm, ISDA~\cite{wang2021regularizing} samples random vectors from a multivariate normal distribution, using the eigenmeans of the source domain as the mean and the covariance matrix for each class conditioned by the source domain. It is possible to discover a wide range of meaningful semantic translation directions by doing so.

Concretely, we begin the analysis by explicitly augmenting each ${\boldsymbol{f}}_{i}$ at $M$ times to form an augmented feature set $\left\{\left(\boldsymbol{f}_{i}^{1}, y_{i}\right), \ldots,\left(\boldsymbol{f}_{i}^{M},y_{i}\right)\right\}_{i=1}^{N}$ of size $MN$, where ${\boldsymbol{f}}_{i}^{m}$ is the $m^{th}$ sample of augmented features for sample $x_i$. With the multi-augmentation, the network for discrimination can be optimized by minimizing the cross-entropy loss as
\begin{equation}
\label{eq:cross-entropy}
\begin{aligned}
\mathcal{L}_{M}(\theta)=\frac{1}{N} \sum_{i=1}^{N} \frac{1}{M} \sum_{n=1}^{M}-\log \left(\frac{e^{\boldsymbol{w}_{y_{i}}^{\mathrm{T}} \boldsymbol{f}_{i}^{m}+b_{y_{i}}}}{\sum_{j=1}^{C} e^{\boldsymbol{w}_{j}^{\mathrm{T}} {\boldsymbol{f}}_{i}^{m}+b_{j}}}\right)
\end{aligned}
\end{equation}	

Then, when $M \rightarrow \infty$,  the CE loss under all possible augmented features can be obtained by the following formulation:
\begin{equation}
\label{eq:infinity}
\begin{aligned}
\mathcal{L}_{\infty}(\theta \mid \boldsymbol{\Sigma})=\frac{1}{N} \sum_{i=1}^{N} \mathrm{E}_{{\boldsymbol{f}}_{i}}\left[-\log \left(\frac{e^{\boldsymbol{w}_{y_{i}}^{\mathrm{T}}} {\boldsymbol{f}}_{i}+b_{y_{i}}}{\sum_{j=1}^{C} e^{\boldsymbol{w}_{j}^{\top} {\boldsymbol{f}}_{i}+b_{j}}}\right)\right]
\end{aligned}
\end{equation}

Then, according to the Jensen's inequality $\mathrm{E}[\log X] \leq \log \mathrm{E}[X]$, as the logarithmic function $\log (\cdot)$, the Eqn.~\ref{eq:infinity} can be written as
\begin{equation}
\label{eq:infinity1}
\begin{aligned}
\mathcal{L}_{\infty}(\theta \mid \boldsymbol{\Sigma})
& = \frac{1}{N} \sum_{i=1}^{N} \log \left(\sum_{j=1}^{C} \mathrm{E}_{{\boldsymbol{f}}_{i}}\left[e^{\boldsymbol{v}_{j y_{i}}^{\mathrm{T}} {\boldsymbol{f}}_{i}+\left(b_{j}-b_{y_{i}}\right)}\right]\right)
\\ &  \leq \frac{1}{N} \sum_{i=1}^{N} \log \left(\sum_{j=1}^{C} \mathrm{E}_{{f}_{i}}\left[e^{\boldsymbol{v}_{j y_{i}}^{\mathrm{T}} {\boldsymbol{f}}_{i}+\left(b_{j}-b_{y_{i}}\right)}\right]\right)
\\ & =\frac{1}{N} \sum_{i=1}^{N} \log (\sum_{j=1}^{C} e^{\boldsymbol{v}_{j y_{i}}^{\mathrm{T}} {\boldsymbol{f}}_{i} + (b_{j}-b_{y_{i}})+\frac{\lambda}{2} \boldsymbol{v}_{j y_{i}}^{\mathrm{T}} \Sigma_{y_{i}} \boldsymbol{v}_{j y_{i}}}).
\end{aligned}
\end{equation}	

\noindent where $\boldsymbol{v}_{j y_{i}}^{\mathrm{T}}=\boldsymbol{w}_{j}^{\mathrm{T}}-\boldsymbol{w}_{y_{i}}^{\mathrm{T}}$,  $\boldsymbol{v}_{j y_{i}}^{\mathrm{T}} {\boldsymbol{f}}_{i} + (b_{j}-b_{y_{i}}) \sim N(v_{jy_{i}}^{\mathrm{T}}\boldsymbol{f}_{i}+(b_{j}-b_{y_{i}}),\lambda v_{jy_{i}}^{\mathrm{T}}\Sigma_{y_{i}}v_{jy_{i}})$, and $\Sigma_{y_{i}}$ comes from ISDA~~\cite{wang2021regularizing}.

The feature augmentation strategy is effective for classification with cross entropy loss. However, the semantic augmentation for DML is more challenging and we will elaborate the strategy to augment features implicitly in the next subsection.

\subsection{Semantic Data Augmentation for DML}
According to the analysis above, the decoupled FC layer makes the semantic augmentation from infinite augmentations feasible. However, many DML losses optimize triplets that only contain original examples. Given a triplet $(x_i, x_j, x_k)$, where $x_i$ and $x_j$ share the same label and $x_k$ is from a different class, DML aims to refine representations such that
\begin{equation}
\label{eq:dml}
\begin{aligned}
\|f(x_i) - f(x_k)\|_2^2 - \|f(x_i) - f(x_j)\|_2^2\geq \delta
\end{aligned}
\end{equation}	
where $\delta$ is a pre-defined margin. Given a set of triplets $\{(x_i^t, x_j^t, x_k^t)\}$, the corresponding loss can be written as
\begin{equation}
\label{eq:dml_hinge}
\begin{aligned}
\mathcal{L}_{DML}(\theta) = \sum_t[\|f(x_i^t) - f(x_j^t)\|_2^2 - \|f(x_i^t) - f(x_k^t)\|_2^2 + \delta ]_+
\end{aligned}
\end{equation}	
where $[]_+$ denotes the hinge loss.

Compared with the cross entropy loss in Eqn.~\ref{eq:infinity}, the coupled features of examples make it hard to apply ISDA for DML directly. To incorporate the semantic augmentation for DML, we first analyze the structure of CE loss. Given the representation of the $i$-th example, the similarity to the $j$-th class is computed as logits in CE loss
\begin{eqnarray}\label{eq:sim}
s_{i,j} = \boldsymbol{w}_{j}^{\mathrm{T}} {\boldsymbol{f}}_{i} + b_{j}
\end{eqnarray}
and the vector $s_i=[s_{i,1},\dots,s_{i,C}]^\top$ denotes the similarity to different classes. According to the analysis in \cite{qian2022kda}, the similarity vector also demonstrates a good approximation for the full pairwise similarity matrix from the perspective of Nystr{\"o}m method~\cite{WilliamsS00}. Inspired by the previous work, we propose to apply logits as features for DML. The distance between pairs of examples can be well bounded by logits as illustrated in the following theorem.
\begin{thm}\label{thm:1}
Let $f_i$ and $s_i$ denote the deep features and similarity vector generated from deep features as in Eqn.~\ref{eq:sim} for the $i$-th example, respectively. Assuming the norm of representations is bounded as $\|f_i\|_2\leq c$, we have
\begin{equation}
\label{eq:dml_have}
\begin{aligned}
& \|s_i-s_j\|_2^2-4c^2 \|UU^\top-\boldsymbol{W} \boldsymbol{W}^\top\|_2 \leq
\\  & \|f_i - f_j\|_2\leq \|s_i - s_j\|_2^2+4c^2 \|UU^\top-\boldsymbol{W} \boldsymbol{W}^\top\|_2
\end{aligned}
\end{equation}	

\end{thm}
\noindent where $U\in R^{d\times k}$ denotes the SVD decomposition for the feature matrix $[f_1,\dots, f_N] = U\Sigma V^\top$.
\begin{proof}
We omit the bias term for brevity. First, with the definition of SVD decomposition, we have
\begin{equation}
\label{eq:dml_have1}
\begin{aligned}
& \|f_i - f_j\|_2^2 = f_i^\top f_i -2f_i^\top f_j+f_j^\top f_j\\
\\ & = f_i^\top (UU^\top- \boldsymbol{W} \boldsymbol{W}^\top) f_i -2f_i^\top (UU^\top- \boldsymbol{W} \boldsymbol{W}^\top) f_j \\
\\ &+ f_j^\top(UU^\top- \boldsymbol{W} \boldsymbol{W}^\top) f_j +f_i^\top\boldsymbol{W} \boldsymbol{W}^\top f_i -2f_i^\top\boldsymbol{W} \boldsymbol{W}^\top f_j+f_j^\top\boldsymbol{W} \boldsymbol{W}^\top f_j\\
\\ & = \|s_i-s_j\|_2^2 + f_i^\top (UU^\top- \boldsymbol{W} \boldsymbol{W}^\top) f_i -2f_i^\top (UU^\top- \boldsymbol{W} \boldsymbol{W}^\top) f_j \\
\\ &+ f_j^\top(UU^\top- \boldsymbol{W} \boldsymbol{W}^\top) f_j
\end{aligned}
\end{equation}	
With the bounded norm for $f_i$ and $f_j$, the desired result can be obtained by Cauchy–Schwarz inequality
\begin{equation}
\label{eq:dml_have2}
\begin{aligned}
\|f_i - f_j\|_2^2\geq\|s_i-s_j\|_2^2-4c^2 \|UU^\top-\boldsymbol{W} \boldsymbol{W}^\top\|_2
\end{aligned}
\end{equation}
and
\begin{equation}
\label{eq:dml_have3}
\begin{aligned}
\|f_i - f_j\|_2^2\leq\|s_i-s_j\|_2^2+4c^2 \|UU^\top-\boldsymbol{W} \boldsymbol{W}^\top\|_2 
\end{aligned}
\end{equation}
\end{proof}

\paragraph{Remark} Note that $W$ denotes the FC layer and contains proxies for different classes. Therefore, an ideal network can have orthogonal class proxies and representations of examples such as the rank $k=C$ and $\|UU^\top-\boldsymbol{W} \boldsymbol{W}^\top\|_2=0$ when $\boldsymbol{w}_j$ has the unit norm. Theorem~\ref{thm:1} implies that the logits $s_i$ can be a good substitute for deep features if the model is optimized appropriately. 

With the logits as features, the loss function for DML can be defined as
\begin{equation}
\label{eq:dml_have4}
\begin{aligned}
\mathcal{L}_{DML}(\theta) = \sum_t[\|s_i^t - s_j^t\|_2^2 - \|s_i^t - s_k^t\|_2^2 + \delta ]_+
\end{aligned}
\end{equation}

Now, we investigate the cross entropy loss with ISDA. By augmenting examples in feature space, CE loss with ISDA can be written as
\begin{equation}
\label{eq:dml_have5}
\begin{aligned}
\mathcal{L}_{CE}^{ISDA} = -\log\frac{e^{\boldsymbol{w}_{y_{i}}^{\mathrm{T}} {\boldsymbol{f}}_{i} + b_{y_{i}}}}{\sum_{j=1}^{C} e^{\boldsymbol{w}_{j}^{\mathrm{T}} {\boldsymbol{f}}_{i} + b_{j}+\frac{\lambda}{2} \boldsymbol{v}_{j y_{i}}^{\mathrm{T}} \Sigma_{y_{i}} \boldsymbol{v}_{j y_{i}}}}
\end{aligned}
\end{equation}

Compared with the conventional CE loss, the logits for the negative class is augmented in the feature space with the covariance matrix. Hence, we have the logits as
\begin{equation}
\label{eq:dml_have6}
\begin{aligned}
s_{i,j}^{ISDA} = \left\{\begin{array}{cc}\boldsymbol{w}_{y_{i}}^{\mathrm{T}} {\boldsymbol{f}}_{i} + b_{y_{i}}&j=i\\\boldsymbol{w}_{j}^{\mathrm{T}} {\boldsymbol{f}}_{i} + b_{j}+\frac{\lambda}{2} \boldsymbol{v}_{j y_{i}}^{\mathrm{T}} \Sigma_{y_{i}} \boldsymbol{v}_{j y_{i}}&o.w.\end{array}\right.
\end{aligned}
\end{equation}
By applying the logits $s_i^{ISDA} = [s_{i,1}^{ISDA},\dots, s_{i,C}^{ISDA}]^\top$ as input features for a DML loss, the examples are augmented with semantic directions for DML implicitly.

\subsection{Final Objective}
With the augmented features, we can define our final objective for effective domain generalization. First, a recent work of Fourier Augmented Co-Teacher (FACT)~\cite{xu2021fourier}, which leverages fourier phase information and amplitude spectrums to improve the generalization ability, is adapted for the DG loss. FACT consists of three-loss including classification loss, Fourier augmented loss and co-teacher regularization loss. Combining all these losses function together, we can get the objective of FACT as
\begin{equation}
\label{eq:loss}
\begin{aligned}
\mathcal{L}_{FACT}=\mathcal{L}_{c l s}^{o r i}+\mathcal{L}_{c l s}^{a u g}+\beta\left(\mathcal{L}_{\text {cot }}^{a 2 o}+\mathcal{L}_{\text {cot }}^{o 2 a}\right)
\end{aligned}
\end{equation}

\noindent where $\beta$ is a trade-off parameter and we set $\beta$ to 2 for Digits-DG and PACS, and 200 for Office-Home which is the same as the original paper FACT.

Then, the lifted structure loss~\cite{oh2016deep} is applied for DML. Given logits with ISDA $\{s_i\}$ as features, it can be defined as

\begin{equation}
\label{eq-03}
\begin{aligned}
& \mathcal{J}_{ij}=\sum_{y_{ij}=1}[\|s_{i}-s_{j}\|_2 + \\&
log(\sum_{y_{ik=0}}exp(m-\|s_{i}-s_{k}\|_2))+log(\sum_{y_{jl}=0}exp(m-\|s_{j}-s_{l}\|_2))]_{+}
\end{aligned}
\end{equation}

\begin{equation}
\label{eq-}
\begin{aligned}
\mathcal{L}_{DML}=\frac{1}{2|\mathcal{P}|} \sum_{(i, j) \in \mathcal{P}} {\mathcal{J}}_{i, j}^{2}
\end{aligned}
\end{equation}
where $y_{ij}=1$ denotes $s_i$ and $s_j$ are from the same class while $y_{ik}=0$ shows that $s_k$ has a different label.

By combining the classification loss and DML loss, our final objective becomes
\begin{equation}
\label{eq-09}
\begin{aligned}
\min _{{\theta}} \mathcal{L}_{FACT}+\alpha \mathcal{L}_{DML}
\end{aligned}
\end{equation}

\noindent where is $\alpha$ is constant controlling the strength of corresponding loss. Note that the logits for DML are defined with a FC layer and we have an additional FC layer for $\mathcal{L}_{DML}$ to model intra-class variance better. By optimizing Eqn.~\ref{eq-09}, clusters of samples belonging to the same category are pulled together in the feature space while synchronously pushing apart from other categories. In this way, our method can simultaneously minimize the domain gap across domains as well as enhance the intra-class compactness and inter-class separability in a unified framework as can be seen in Fig.~\ref{fig:framework}.

\section{Experiments}

Our method is compared to three benchmarks for DG in this section. Additionally, we conduct detailed ablation studies to examine how different components affect the results.

\subsection{Data Description}

\noindent \textbf{Digits-DG}~\cite{zhou2020deep} is a digit recognition benchmark which consists of four digit datasets: MNIST~\cite{lecun1998gradient},MNIST-M~\cite{ganin2015unsupervised},SVHN~\cite{netzer2011reading} and SYN~\cite{ganin2015unsupervised}. The four datasets mainly contain four different image quality, font style and background. MNIST~\cite{lecun1998gradient} contains 10 categories of handwritten digits dataset. MNIST-M~\cite{ganin2015unsupervised} is a variant of MNIST~\cite{lecun1998gradient} by blending the image with random color patches. SVHN~\cite{netzer2011reading} contains street view house number images. SYN~\cite{ganin2015unsupervised} consists of synthetic digital images with different fonts, backgrounds and stroke colors.

\noindent \textbf{PACS}~\cite{li2017deeper} is consisted of four domains, namely Photo (1,670 images), Art(2,048 images) Cartoon(2,344 images) and Sketch(3,929 images). Each domain contains seven categories. Following the previous work~\cite{li2017deeper}, we choose one domain as the test domain and use the remaining three domains as the source domains. 
For a fair comparison with the published methods, our model is trained using only data from the training split.

\noindent \textbf{Office-Home}~\cite{venkateswara2017deep} is a more complex dataset than PACS~\cite{li2017deeper} , which contains a total of 15,500 images in 65 categories . There exist four extremely different domains in the Office-Home dataset: Artistic images, Product images, Clipart images and Real-World images. 

\subsection{Implementation Details}
 We leverage a popular model (FACT) as our baseline to test the effectiveness of our method. We also compare our proposed method with a variety of state-of-the-art domain generalization methods, including DeepAll~\cite{zhou2020deep}, Jigen~\cite{carlucci2019domain}, CCSA~\cite{motiian2017unified}, MMD-AAE~\cite{li2018domain} , CrossGrad~\cite{shankar2018generalizing}, DDAIG~\cite{zhou2020learning}, L2A-OT~\cite{zhou2020deep}, ATSRL~\cite{yang2021adversarial}, MetaReg~\cite{balaji2018metareg} , Epi-FCR~\cite{li2019episodic}, MMLD~\cite{matsuura2020domain}, CSD~\cite{piratla2020efficient}, InfoDrop~\cite{shi2020informative}, MASF~\cite{dou2019domain}, Mixstyle~\cite{zhou2021domain}, EISNet~\cite{wang2020learning}, MDGH~\cite{mahajan2021domain}, RSC~\cite{huang2020self} and FACT~\cite{xu2021fourier}. The results of baselines are cited from the original papers.

\subsection{Results on Digits-DG}
The results are presented in Table.~\ref{digitsdg}. Our method is the best among all competitors, beating the second-best method FACT~\cite{xu2021fourier} on average by 2\%. Our method outperforms FACT with a large margin of 1.9\% and 2.2\%, respectively, on the hardest target domains SVHN and SYN, where cluttered digits and poor image quality are involved. According to the results of our method, logits with semantic augmentation can improve the DML loss' performance when used as the input features for replace the original deep features on leave-one-domain-out images.

\begin{table}[!t]
\caption{Digits-DG results with a domain left out. We have bolded and underlined the best and second-best results.}
  \centering
    \setlength{\tabcolsep}{0.8mm}{\begin{tabular}{l|cccc|c}
    \toprule
    Methods & MNIST & MNIST-M & SVHN & SYN & Avg. \\
    \midrule
    DeepAll~\cite{zhou2020deep} & 95.8 & 58.8 & 61.7 & 78.6 & 73.7 \\
    Jigen~\cite{carlucci2019domain} & 96.5& 61.4 & 63.7 & 74.0 & 73.9 \\
    CCSA~\cite{motiian2017unified} & 95.2 & 58.2 & 65.5 & 79.1 & 74.5 \\
    MMD-AAE~\cite{li2018domain} & 96.5 & 58.4 & 65.0 & 78.4 & 74.6 \\
    CrossGrad~\cite{shankar2018generalizing} & 96.7 & 61.1 & 65.3 & 80.2 & 75.8 \\
    DDAIG~\cite{zhou2020learning} & 96.6 & 64.1 & 68.6 & 81.0 & 77.6 \\
    L2A-OT~\cite{zhou2020deep} & 96.7 & 63.9 & 68.6 & 83.2 & 78.1 \\
    ATSRL~\cite{yang2021adversarial} & \underline{97.9} & 62.7 & 69.3 & 83.7 & 78.4 \\
    \hline
    FACT (\textit{baseline})~\cite{xu2021fourier} & \underline{97.9} & \underline{65.6} & \underline{72.4} & \underline{90.3} & \underline{81.5} \\
    
    Ours & \textbf{98.9} & \textbf{68.2} & \textbf{74.3} & \textbf{92.5} & \textbf{83.5} \\
    \bottomrule
    \end{tabular}}
  \label{digitsdg}
\end{table}

\begin{table}[!t]
  \centering
  \caption{PACS results with a domain left out. We have bolded and underlined the best and second-best results.}
    \setlength{\tabcolsep}{1.0mm}{\begin{tabular}{l|cccc|c}
    \toprule
    Methods & Art & Cartoon & Photo & Sketch & Avg. \\
    \midrule
    \multicolumn{6}{c}{\textit{ResNet18}} \\
    \midrule
    DeepAll~\cite{zhou2020deep} & 77.63 & 76.77 & 95.85 & 69.50 & 79.94 \\
    MetaReg~\cite{balaji2018metareg} & 83.70& 77.20 & 95.50 & 70.30 & 81.70 \\
    CrossGrad~\cite{shankar2018generalizing} & 79.8 & 76.8 & 96.0 & 70.2 & 80.70 \\
    JiGen~\cite{carlucci2019domain} & 79.42 & 75.25 & 96.03 & 71.35 & 80.51 \\
    Epi-FCR~\cite{li2019episodic} & 82.10 & 77.00 & 93.90 & 73.00 & 81.50 \\
    MMLD~\cite{matsuura2020domain}  & 81.28 & 77.16 & 96.09 & 72.29 & 81.83 \\
    DDAIG~\cite{zhou2020learning} & 84.20 & 78.10 & 95.30 & 74.70 & 83.10 \\
    CSD~\cite{piratla2020efficient} & 78.90 & 75.80 & 94.10 & 76.70 & 81.40 \\
    InfoDrop~\cite{shi2020informative} & 80.27 & 76.54 & 96.11 & 76.38 & 82.33 \\
    MASF~\cite{dou2019domain} & 80.29 & 77.17 & 94.99 & 71.69 & 81.04 \\
    L2A-OT~\cite{zhou2020deep} & 83.30 & 78.20 & \underline{96.20} & 73.60 & 82.80 \\
    MixStyle~\cite{zhou2021domain} & 84.1 & 78.8 & 96.1 & 75.9 & 83.7 \\
    EISNet~\cite{wang2020learning} & 81.89 & 76.44 & 95.93 & 74.33 & 82.15 \\
    \hline
    FACT (\textit{baseline})~\cite{xu2021fourier}  & \underline{85.37} & \underline{78.38} & 95.15 & \underline{79.15} & \underline{84.51} \\
    Ours  & \textbf{87.96} & \textbf{82.41} & \textbf{98.85} & \textbf{83.21} & \textbf{88.08} \\
    
    \midrule
    \multicolumn{6}{c}{\textit{ResNet50}} \\
    \midrule
    DeepAll~\cite{zhou2020deep} & 84.94 & 76.98 & 97.64 & 76.75 & 84.08 \\
    MetaReg~\cite{balaji2018metareg} & 87.20 & 79.20 & 97.60 & 70.30 & 83.60 \\
    DDAIG~\cite{zhou2020learning} & 85.4 & 78.5 & 95.7 & 80.0 & 84.9 \\
    CrossGrad~\cite{shankar2018generalizing} & 87.5 & 80.7 & \underline{97.8} & 73.9 & 85.7 \\
    MASF~\cite{dou2019domain}  & 82.89 & 80.49 & 95.01 & 72.29 & 82.67 \\
    EISNet~\cite{wang2020learning} & 86.64 & 81.53 & 97.11 & 78.07 & 85.84 \\
    RSC~\cite{huang2020self} & 87.89 & 82.16 & 97.92 & 83.35 & 87.83 \\
    ATSRL~\cite{yang2021adversarial}  & \underline{90.0} & \underline{83.5} & 98.9 & 80.0 & 88.1 \\
    MDGH~\cite{mahajan2021domain} & 86.7 & 82.3 & 98.4 & 82.7 & 87.5 \\
    \hline
    FACT (\textit{baseline})~\cite{xu2021fourier}  & 89.63 & 81.77 & 96.75 & \underline{84.46} & \underline{88.15} \\
    Ours & \textbf{92.43} & \textbf{84.94} & \textbf{99.13} & \textbf{88.29} & \textbf{91.20} \\
    \bottomrule
    \end{tabular}}
  \label{pacs}
\end{table}

\begin{table}[!t]
  \centering
  \caption{Office-Home results with a domain left out. We have bolded and underlined the best and second-best results.}
    \setlength{\tabcolsep}{1.0mm}{\begin{tabular}{l|cccc|c}
    \toprule
    Methods & Art & Clipart & Product & Real & Avg. \\
    \midrule
    DeepAll~\cite{zhou2020deep} & 57.88 & 52.72 & 73.50 & 74.80 & 64.72 \\
    CCSA~\cite{motiian2017unified}  & 59.90 & 49.90 & 74.10 & 75.70 & 64.90 \\
    MMD-AAE~\cite{li2018domain} & 56.50 & 47.30 & 72.10 & 74.80 & 62.70 \\
    CrossGrad~\cite{shankar2018generalizing} & 58.40 & 49.40 & 73.90 & 75.80 & 64.40 \\
    DDAIG~\cite{zhou2020learning} & 59.20 & 52.30 & 74.60 & 76.00 & 65.50 \\
    L2A-OT~\cite{zhou2020deep} & \underline{60.60} & 50.10 & \underline{74.80} & \underline{77.00} &65.60 \\
    Jigen~\cite{carlucci2019domain} & 53.04 & 47.51 & 71.47 & 72.79 & 61.20 \\
    RSC~\cite{huang2020self}  & 58.42 & 47.90 & 71.63 & 74.54 & 63.12 \\
    \hline
    FACT (\textit{baseline})~\cite{xu2021fourier} & 60.34 & \underline{54.85} & 74.48 & 76.55 & \underline{66.56} \\
    Ours & \textbf{68.29} & \textbf{57.63} & \textbf{76.21} & \textbf{80.69} & \textbf{70.71} \\
    \bottomrule
    \end{tabular}}
  \label{officehome}
\end{table}

\subsection{Results on PACS}
The results of PACS are shown in Table.~\ref{pacs}. It can be seen that our method achieves the best results and outperforms the existing methods by a large margin. Meanwhile, we notice that our method also performs well on the large domain discrepancy task cartoon and sketch domains. We also notice that our method lifts a tremendous margin of 4.06\% on ResNet-18 and 3.83\% on ResNet-50, respectively. Meanwhile, the performance of our method on the photo domain improved performance is 
also better than other domains. We believe our method captures meaningful information in domains like Photo, since these details are redundant and complex.The current state-of-the-art methods clearly outperform ours, which is a result of the GAN-based augmentation method and meta-learning. Furthermore, our method does not require additional adversarial training due to its efficient training process. We have shown through all these comparisons how effective our method is, and that it is simpler and more efficient than other methods.

\subsection{Results on Office-Home}
Our method acts as a good baseline for Table~\ref{officehome} due to its relatively smaller domain discrepancy and similarity to ImageNet's pretrained dataset. Many previous DG methods, such as CCSA~\cite{motiian2017unified} , MMD-AAE~\cite{li2018domain}, CrossGrad~\cite{shankar2018generalizing} and Jigen~\cite{zhou2020deep}, can not improve much over the tasks. Nevertheless, our method achieves a consistent improvement over all the leave-one-domain-out domains. Further, we achieve better average performance than the latest versions of RSC~\cite{huang2020self} and L2A-OT~\cite{zhou2020deep}.Hence, our method is superior.

\section{Analysis}

\begin{table}[!t]
  \centering
  \caption{The Office-Home dataset was used for ablation studies of different elements of our method.}
    \setlength{\tabcolsep}{1.0mm}{\begin{tabular}{l|cccc|c}
    \toprule
    Methods & Art & Clipart & Product & Real & Avg. \\
    \midrule
    \hline
    FACT (\textit{baseline})~\cite{xu2021fourier} & 60.34 & 54.85 & 74.48 & 76.55 & 66.56 \\
    FACT (\textit{+DML (features)})~\cite{oh2016deep} & 63.58 & 55.61 & 74.73 & 78.02 & 67.99\\ 
    FACT (\textit{+DML (logits)})~\cite{oh2016deep} & 64.28 & 56.21 & 75.39 & 78.94 & 68.71\\
    FACT (\textit{+ISDA})~\cite{wang2021regularizing} & 63.27 & 55.31 & 75.67 & 77.77 & 68.00 \\ 
    Ours & \textbf{68.29} & \textbf{57.63} & \textbf{76.21} & \textbf{80.69} & \textbf{70.71} \\
    \bottomrule
    \end{tabular}}
  \label{ablation study}
\end{table}

\begin{table}[!t]
  \centering
  \caption{Results of our method combine other baselines with ResNet18.}
    \setlength{\tabcolsep}{1.0mm}{\begin{tabular}{l|cccc|c}
    \toprule
    Methods & Art & Cartoon & Photo & Sketch & Avg. \\
    \midrule
    \hline
    DG\_via\_ER~\cite{zhao2020domain} & 80.70 & 76.40 & 96.65 & 71.77 & 81.38 \\ 
    DG\_via\_ER+Ours & \textbf{83.14} & \textbf{80.07} & \textbf{98.35} & \textbf{74.30} & \textbf{83.96} \\ 
    \hline
    JiGen~\cite{carlucci2019domain} & 79.42 & 75.25 & 96.03 & 71.35 & 80.51 \\  
    JiGen+Ours & \textbf{82.37} & \textbf{76.94} & \textbf{97.84} & \textbf{73.39} & \textbf{82.64} \\ 
    \hline
    EISNet~\cite{wang2020learning} & 81.89 & 76.44 & 75.93 & 74.33 & 82.15 \\  
    EISNet+Ours & \textbf{83.33} & \textbf{78.69} & \textbf{76.48} & \textbf{76.38} & \textbf{83.72} \\     
    \bottomrule
    \end{tabular}}
  \label{effectiveness}
\end{table}

\begin{figure*}[t]
	\centering
	\includegraphics[width=0.85\linewidth]{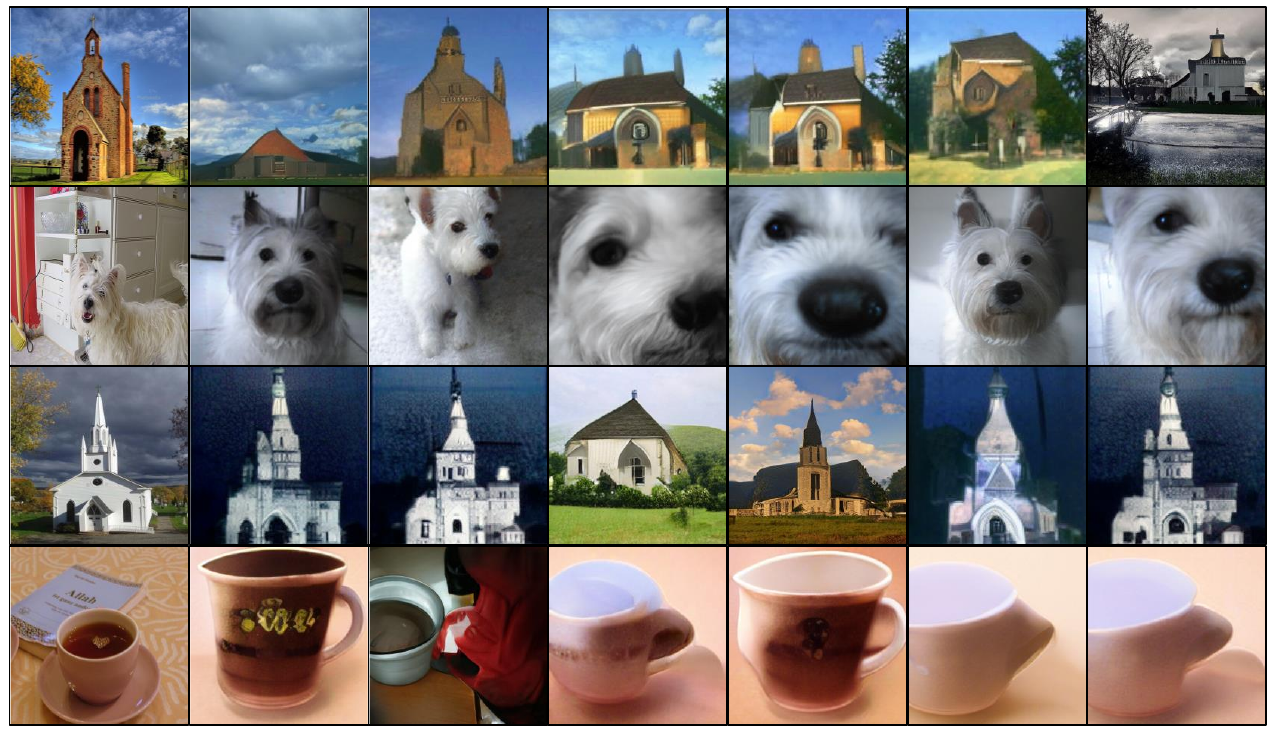}
	\caption{Visualization of the semantically augmented images. Images that do not pertain to the identity of the class can be re-semantized by our method, such as backgrounds, animal actions, visual angles, etc.} 
	\label{fig:fig3}
\end{figure*}
\noindent \textbf{Ablation Study.} The ablation study in Table.~\ref{ablation study} investigates the role of the various components in our method. FACT (\textit{baseline}) denotes that we only use FACT without any other strategy. Based on FACT (\textit{baseline}), we add a DML loss~\cite{oh2016deep} to obtain  FACT (\textit{+DML})~\cite{oh2016deep} (here we leverage original features and logits as the inputs, which denotes FACT (+ DML (features)) and FACT (+DML (logits))), which improves over FACT (\textit{baseline}) slightly. Further incorporating the implicit semantic data augmentation (ISDA), which perform better than FACT (\textit{+DML})~\cite{oh2016deep}. This indicates that leverage logits to replace features can provide better domain generalization ability. \\

\noindent \textbf{Effectiveness of our method.} To verify the effectiveness of our proposed method, we tested it on three other baselines including DG\_via\_ER~\cite{zhao2020domain}, JiGen~\cite{carlucci2019domain} and EISNet~\cite{wang2020learning} in Table.~\ref{effectiveness}, we find that JiGen+Ours,  DG\_via\_ER+Ours, EISNet+Ours improve the baseline 2.13\%, 2.58\% and 1.57\%, respectively. This not only demonstrates the effectiveness of our proposed method, but also shows that our method is a plug-and-play method. \\

\noindent \textbf{Visualization of augmented images.} We present an approach to map the augmented features back to the original images in order to illustrate that our method can generate meaningful semantically augmented samples.  Figure.~\ref{fig:fig3} shows the results of the visualization. Original images are displayed in the first column. The other columns present the images augmented by our method. With the use of our technique, we were able to alter the semantics of images, for example, backgrounds, visual angles, the actions of dogs, and skin colors, which is not possible with traditional data augmentation techniques. 

\begin{figure}[tb]
	\centering
	\includegraphics[width=\linewidth]{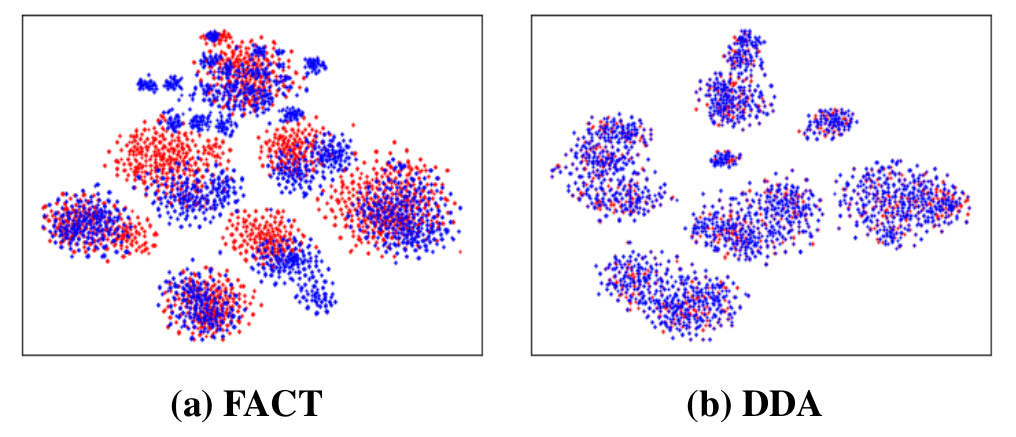}
	\caption{FACT and Ours' deep features visualized on PACS. (Target: Photo)} 
	\label{fig:tsne}
\end{figure}

\noindent \textbf{Visualization of Deep Features.} We visualize the learned deep features on  PACS using the t-SNE algorithm~\cite{van2008visualizing} in Fig.~\ref{fig:tsne}. We use "Photo" as the target domain and "Art-Cartoon-Sketch" as the source domain. From the t-SNE~\cite{van2008visualizing} analysis, we can observe that previous FACT category alignment method could produce separated features, yet it may be hard for dense prediction since the margins between different category features are not obvious and the distribution is still dispersed. When we use our method, features among different categories are better separated, demonstrating that the semantic distributions can provide correct supervision signal for unknown target domain. In comparison, the embedded representations of our our method exhibit the clearest clusters compared with other state-of-the-arts, revealing the discriminative capability of the domain generalization. \\

\begin{figure}[tb]
	\centering
	\includegraphics[width=\linewidth]{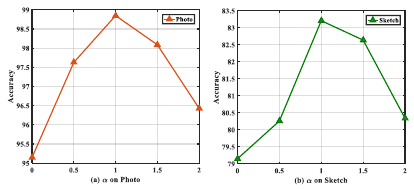}
	\caption{Sensitivity analysis of our method.} 
	\label{fig5}
\end{figure}

\noindent \textbf{Parameter Sensitivity.} To study how the hyper-parameter $\alpha$ affects the performance of our method, sensitivity tests are conducted for PACS on task Photo and Sketch. The results are shown in Fig.\ref{fig5}. When $\alpha=0$, there is only the baseline of FACT. From Fig.\ref{fig5},  we can see that when $\alpha=1$, the results for the datasets of PACS are the best.

\section{Conclusion}
In this paper, we propose a semantic data augmentation based distance metric learning for domain generalization. We leverage a novel implicit semantic augmentation strategy for DML loss in domain generalization. We find that the logits with the semantic augmentation to replace the input features for DML loss is more suitable for domain generalization. We also provide a theoretical analysis to show that the logits can approximate the distances defined on original features well. Lastly, we provide an in-depth analysis of the rationale and mechanism behind our approach, enabling us to better understand why using logits instead of features can help generalize domains. The results of extensive experiments on three benchmarks demonstrate our method achieves world-class performance in domain generalization. In light of the fact that domain-adversarial learning still dominates the field, we hope our work can provide some inspiration for the community.

\bibliographystyle{ACM-Reference-Format}
\balance
\bibliography{sample-sigconf}


\begin{thebibliography}{60}


\ifx \showCODEN    \undefined \def \showCODEN     #1{\unskip}     \fi
\ifx \showDOI      \undefined \def \showDOI       #1{#1}\fi
\ifx \showISBNx    \undefined \def \showISBNx     #1{\unskip}     \fi
\ifx \showISBNxiii \undefined \def \showISBNxiii  #1{\unskip}     \fi
\ifx \showISSN     \undefined \def \showISSN      #1{\unskip}     \fi
\ifx \showLCCN     \undefined \def \showLCCN      #1{\unskip}     \fi
\ifx \shownote     \undefined \def \shownote      #1{#1}          \fi
\ifx \showarticletitle \undefined \def \showarticletitle #1{#1}   \fi
\ifx \showURL      \undefined \def \showURL       {\relax}        \fi
\providecommand\bibfield[2]{#2}
\providecommand\bibinfo[2]{#2}
\providecommand\natexlab[1]{#1}
\providecommand\showeprint[2][]{arXiv:#2}

\bibitem[Ao et~al\mbox{.}(2017)]%
        {ao2017fast}
\bibfield{author}{\bibinfo{person}{Shuang Ao}, \bibinfo{person}{Xiang Li},
  {and} \bibinfo{person}{Charles Ling}.} \bibinfo{year}{2017}\natexlab{}.
\newblock \showarticletitle{Fast generalized distillation for semi-supervised
  domain adaptation}. In \bibinfo{booktitle}{\emph{Proceedings of the AAAI
  Conference on Artificial Intelligence}}, Vol.~\bibinfo{volume}{31}.
\newblock


\bibitem[Balaji et~al\mbox{.}(2018)]%
        {balaji2018metareg}
\bibfield{author}{\bibinfo{person}{Yogesh Balaji}, \bibinfo{person}{Swami
  Sankaranarayanan}, {and} \bibinfo{person}{Rama Chellappa}.}
  \bibinfo{year}{2018}\natexlab{}.
\newblock \showarticletitle{Metareg: Towards domain generalization using
  meta-regularization}.
\newblock \bibinfo{journal}{\emph{Advances in neural information processing
  systems}}  \bibinfo{volume}{31} (\bibinfo{year}{2018}).
\newblock


\bibitem[Carlucci et~al\mbox{.}(2019)]%
        {carlucci2019domain}
\bibfield{author}{\bibinfo{person}{Fabio~M Carlucci}, \bibinfo{person}{Antonio
  D'Innocente}, \bibinfo{person}{Silvia Bucci}, \bibinfo{person}{Barbara
  Caputo}, {and} \bibinfo{person}{Tatiana Tommasi}.}
  \bibinfo{year}{2019}\natexlab{}.
\newblock \showarticletitle{Domain generalization by solving jigsaw puzzles}.
  In \bibinfo{booktitle}{\emph{Proceedings of the IEEE/CVF Conference on
  Computer Vision and Pattern Recognition}}. \bibinfo{pages}{2229--2238}.
\newblock


\bibitem[Chen et~al\mbox{.}(2019)]%
        {chen2019progressive}
\bibfield{author}{\bibinfo{person}{Chaoqi Chen}, \bibinfo{person}{Weiping Xie},
  \bibinfo{person}{Wenbing Huang}, \bibinfo{person}{Yu Rong},
  \bibinfo{person}{Xinghao Ding}, \bibinfo{person}{Yue Huang},
  \bibinfo{person}{Tingyang Xu}, {and} \bibinfo{person}{Junzhou Huang}.}
  \bibinfo{year}{2019}\natexlab{}.
\newblock \showarticletitle{Progressive feature alignment for unsupervised
  domain adaptation}. In \bibinfo{booktitle}{\emph{Proceedings of the IEEE/CVF
  Conference on Computer Vision and Pattern Recognition}}.
  \bibinfo{pages}{627--636}.
\newblock


\bibitem[Cui et~al\mbox{.}(2020)]%
        {cui2020towards}
\bibfield{author}{\bibinfo{person}{Shuhao Cui}, \bibinfo{person}{Shuhui Wang},
  \bibinfo{person}{Junbao Zhuo}, \bibinfo{person}{Liang Li},
  \bibinfo{person}{Qingming Huang}, {and} \bibinfo{person}{Qi Tian}.}
  \bibinfo{year}{2020}\natexlab{}.
\newblock \showarticletitle{Towards discriminability and diversity: Batch
  nuclear-norm maximization under label insufficient situations}. In
  \bibinfo{booktitle}{\emph{Proceedings of the IEEE/CVF Conference on Computer
  Vision and Pattern Recognition}}. \bibinfo{pages}{3941--3950}.
\newblock


\bibitem[Donahue et~al\mbox{.}(2013)]%
        {donahue2013semi}
\bibfield{author}{\bibinfo{person}{Jeff Donahue}, \bibinfo{person}{Judy
  Hoffman}, \bibinfo{person}{Erik Rodner}, \bibinfo{person}{Kate Saenko}, {and}
  \bibinfo{person}{Trevor Darrell}.} \bibinfo{year}{2013}\natexlab{}.
\newblock \showarticletitle{Semi-supervised domain adaptation with instance
  constraints}. In \bibinfo{booktitle}{\emph{Proceedings of the IEEE conference
  on computer vision and pattern recognition}}. \bibinfo{pages}{668--675}.
\newblock


\bibitem[Dou et~al\mbox{.}(2019)]%
        {dou2019domain}
\bibfield{author}{\bibinfo{person}{Qi Dou}, \bibinfo{person}{Daniel Coelho~de
  Castro}, \bibinfo{person}{Konstantinos Kamnitsas}, {and} \bibinfo{person}{Ben
  Glocker}.} \bibinfo{year}{2019}\natexlab{}.
\newblock \showarticletitle{Domain generalization via model-agnostic learning
  of semantic features}.
\newblock \bibinfo{journal}{\emph{Advances in Neural Information Processing
  Systems}}  \bibinfo{volume}{32} (\bibinfo{year}{2019}).
\newblock


\bibitem[Ganin and Lempitsky(2015)]%
        {ganin2015unsupervised}
\bibfield{author}{\bibinfo{person}{Yaroslav Ganin} {and}
  \bibinfo{person}{Victor Lempitsky}.} \bibinfo{year}{2015}\natexlab{}.
\newblock \showarticletitle{Unsupervised domain adaptation by backpropagation}.
  In \bibinfo{booktitle}{\emph{International conference on machine learning}}.
  \bibinfo{pages}{1180--1189}.
\newblock


\bibitem[Ghifary et~al\mbox{.}(2016)]%
        {ghifary2016scatter}
\bibfield{author}{\bibinfo{person}{Muhammad Ghifary}, \bibinfo{person}{David
  Balduzzi}, \bibinfo{person}{W~Bastiaan Kleijn}, {and}
  \bibinfo{person}{Mengjie Zhang}.} \bibinfo{year}{2016}\natexlab{}.
\newblock \showarticletitle{Scatter component analysis: A unified framework for
  domain adaptation and domain generalization}.
\newblock \bibinfo{journal}{\emph{IEEE transactions on pattern analysis and
  machine intelligence}} (\bibinfo{year}{2016}), \bibinfo{pages}{1414--1430}.
\newblock


\bibitem[Hoffman et~al\mbox{.}(2018)]%
        {hoffman2018cycada}
\bibfield{author}{\bibinfo{person}{Judy Hoffman}, \bibinfo{person}{Eric Tzeng},
  \bibinfo{person}{Taesung Park}, \bibinfo{person}{Jun-Yan Zhu},
  \bibinfo{person}{Phillip Isola}, \bibinfo{person}{Kate Saenko},
  \bibinfo{person}{Alexei Efros}, {and} \bibinfo{person}{Trevor Darrell}.}
  \bibinfo{year}{2018}\natexlab{}.
\newblock \showarticletitle{Cycada: Cycle-consistent adversarial domain
  adaptation}. In \bibinfo{booktitle}{\emph{International conference on machine
  learning}}. \bibinfo{pages}{1989--1998}.
\newblock


\bibitem[Huang et~al\mbox{.}(2020)]%
        {huang2020self}
\bibfield{author}{\bibinfo{person}{Zeyi Huang}, \bibinfo{person}{Haohan Wang},
  \bibinfo{person}{Eric~P Xing}, {and} \bibinfo{person}{Dong Huang}.}
  \bibinfo{year}{2020}\natexlab{}.
\newblock \showarticletitle{Self-challenging improves cross-domain
  generalization}. In \bibinfo{booktitle}{\emph{European Conference on Computer
  Vision}}. \bibinfo{pages}{124--140}.
\newblock


\bibitem[Kim and Byun(2020)]%
        {kim2020learning}
\bibfield{author}{\bibinfo{person}{Myeongjin Kim} {and} \bibinfo{person}{Hyeran
  Byun}.} \bibinfo{year}{2020}\natexlab{}.
\newblock \showarticletitle{Learning texture invariant representation for
  domain adaptation of semantic segmentation}. In
  \bibinfo{booktitle}{\emph{Proceedings of the IEEE/CVF conference on computer
  vision and pattern recognition}}. \bibinfo{pages}{12975--12984}.
\newblock


\bibitem[LeCun et~al\mbox{.}(1998)]%
        {lecun1998gradient}
\bibfield{author}{\bibinfo{person}{Yann LeCun}, \bibinfo{person}{L{\'e}on
  Bottou}, \bibinfo{person}{Yoshua Bengio}, {and} \bibinfo{person}{Patrick
  Haffner}.} \bibinfo{year}{1998}\natexlab{}.
\newblock \showarticletitle{Gradient-based learning applied to document
  recognition}.
\newblock \bibinfo{journal}{\emph{Proc. IEEE}} (\bibinfo{year}{1998}),
  \bibinfo{pages}{2278--2324}.
\newblock


\bibitem[Li et~al\mbox{.}(2017)]%
        {li2017deeper}
\bibfield{author}{\bibinfo{person}{Da Li}, \bibinfo{person}{Yongxin Yang},
  \bibinfo{person}{Yi-Zhe Song}, {and} \bibinfo{person}{Timothy~M Hospedales}.}
  \bibinfo{year}{2017}\natexlab{}.
\newblock \showarticletitle{Deeper, broader and artier domain generalization}.
  In \bibinfo{booktitle}{\emph{Proceedings of the IEEE international conference
  on computer vision}}. \bibinfo{pages}{5542--5550}.
\newblock


\bibitem[Li et~al\mbox{.}(2019)]%
        {li2019episodic}
\bibfield{author}{\bibinfo{person}{Da Li}, \bibinfo{person}{Jianshu Zhang},
  \bibinfo{person}{Yongxin Yang}, \bibinfo{person}{Cong Liu},
  \bibinfo{person}{Yi-Zhe Song}, {and} \bibinfo{person}{Timothy~M Hospedales}.}
  \bibinfo{year}{2019}\natexlab{}.
\newblock \showarticletitle{Episodic training for domain generalization}. In
  \bibinfo{booktitle}{\emph{Proceedings of the IEEE/CVF International
  Conference on Computer Vision}}. \bibinfo{pages}{1446--1455}.
\newblock


\bibitem[Li et~al\mbox{.}(2018a)]%
        {li2018domain}
\bibfield{author}{\bibinfo{person}{Haoliang Li}, \bibinfo{person}{Sinno~Jialin
  Pan}, \bibinfo{person}{Shiqi Wang}, {and} \bibinfo{person}{Alex~C Kot}.}
  \bibinfo{year}{2018}\natexlab{a}.
\newblock \showarticletitle{Domain generalization with adversarial feature
  learning}. In \bibinfo{booktitle}{\emph{Proceedings of the IEEE conference on
  computer vision and pattern recognition}}. \bibinfo{pages}{5400--5409}.
\newblock


\bibitem[Li et~al\mbox{.}(2021b)]%
        {li2021semantic}
\bibfield{author}{\bibinfo{person}{Shuang Li}, \bibinfo{person}{Binhui Xie},
  \bibinfo{person}{Bin Zang}, \bibinfo{person}{Chi~Harold Liu},
  \bibinfo{person}{Xinjing Cheng}, \bibinfo{person}{Ruigang Yang}, {and}
  \bibinfo{person}{Guoren Wang}.} \bibinfo{year}{2021}\natexlab{b}.
\newblock \showarticletitle{Semantic distribution-aware contrastive adaptation
  for semantic segmentation}.
\newblock \bibinfo{journal}{\emph{arXiv preprint arXiv:2105.05013}}
  (\bibinfo{year}{2021}).
\newblock


\bibitem[Li et~al\mbox{.}(2021a)]%
        {li2021transferable}
\bibfield{author}{\bibinfo{person}{Shuang Li}, \bibinfo{person}{Mixue Xie},
  \bibinfo{person}{Kaixiong Gong}, \bibinfo{person}{Chi~Harold Liu},
  \bibinfo{person}{Yulin Wang}, {and} \bibinfo{person}{Wei Li}.}
  \bibinfo{year}{2021}\natexlab{a}.
\newblock \showarticletitle{Transferable semantic augmentation for domain
  adaptation}. In \bibinfo{booktitle}{\emph{Proceedings of the IEEE/CVF
  Conference on Computer Vision and Pattern Recognition}}.
  \bibinfo{pages}{11516--11525}.
\newblock


\bibitem[Li et~al\mbox{.}(2018b)]%
        {li2018deep}
\bibfield{author}{\bibinfo{person}{Ya Li}, \bibinfo{person}{Xinmei Tian},
  \bibinfo{person}{Mingming Gong}, \bibinfo{person}{Yajing Liu},
  \bibinfo{person}{Tongliang Liu}, \bibinfo{person}{Kun Zhang}, {and}
  \bibinfo{person}{Dacheng Tao}.} \bibinfo{year}{2018}\natexlab{b}.
\newblock \showarticletitle{Deep domain generalization via conditional
  invariant adversarial networks}. In \bibinfo{booktitle}{\emph{Proceedings of
  the European Conference on Computer Vision (ECCV)}}.
  \bibinfo{pages}{624--639}.
\newblock


\bibitem[Mahajan et~al\mbox{.}(2021)]%
        {mahajan2021domain}
\bibfield{author}{\bibinfo{person}{Divyat Mahajan}, \bibinfo{person}{Shruti
  Tople}, {and} \bibinfo{person}{Amit Sharma}.}
  \bibinfo{year}{2021}\natexlab{}.
\newblock \showarticletitle{Domain generalization using causal matching}. In
  \bibinfo{booktitle}{\emph{International Conference on Machine Learning}}.
  \bibinfo{pages}{7313--7324}.
\newblock


\bibitem[Matsuura and Harada(2020)]%
        {matsuura2020domain}
\bibfield{author}{\bibinfo{person}{Toshihiko Matsuura} {and}
  \bibinfo{person}{Tatsuya Harada}.} \bibinfo{year}{2020}\natexlab{}.
\newblock \showarticletitle{Domain generalization using a mixture of multiple
  latent domains}. In \bibinfo{booktitle}{\emph{Proceedings of the AAAI
  Conference on Artificial Intelligence}}. \bibinfo{pages}{11749--11756}.
\newblock


\bibitem[Montavon et~al\mbox{.}(2018)]%
        {montavon2018methods}
\bibfield{author}{\bibinfo{person}{Gr{\'e}goire Montavon},
  \bibinfo{person}{Wojciech Samek}, {and} \bibinfo{person}{Klaus-Robert
  M{\"u}ller}.} \bibinfo{year}{2018}\natexlab{}.
\newblock \showarticletitle{Methods for interpreting and understanding deep
  neural networks}.
\newblock \bibinfo{journal}{\emph{Digital Signal Processing}}
  (\bibinfo{year}{2018}), \bibinfo{pages}{1--15}.
\newblock


\bibitem[Motiian et~al\mbox{.}(2017)]%
        {motiian2017unified}
\bibfield{author}{\bibinfo{person}{Saeid Motiian}, \bibinfo{person}{Marco
  Piccirilli}, \bibinfo{person}{Donald~A Adjeroh}, {and}
  \bibinfo{person}{Gianfranco Doretto}.} \bibinfo{year}{2017}\natexlab{}.
\newblock \showarticletitle{Unified deep supervised domain adaptation and
  generalization}. In \bibinfo{booktitle}{\emph{Proceedings of the IEEE
  international conference on computer vision}}. \bibinfo{pages}{5715--5725}.
\newblock


\bibitem[Muandet et~al\mbox{.}(2013)]%
        {muandet2013domain}
\bibfield{author}{\bibinfo{person}{Krikamol Muandet}, \bibinfo{person}{David
  Balduzzi}, {and} \bibinfo{person}{Bernhard Sch{\"o}lkopf}.}
  \bibinfo{year}{2013}\natexlab{}.
\newblock \showarticletitle{Domain generalization via invariant feature
  representation}. In \bibinfo{booktitle}{\emph{International Conference on
  Machine Learning}}. \bibinfo{pages}{10--18}.
\newblock


\bibitem[Netzer et~al\mbox{.}(2011)]%
        {netzer2011reading}
\bibfield{author}{\bibinfo{person}{Yuval Netzer}, \bibinfo{person}{Tao Wang},
  \bibinfo{person}{Adam Coates}, \bibinfo{person}{Alessandro Bissacco},
  \bibinfo{person}{Bo Wu}, {and} \bibinfo{person}{Andrew~Y Ng}.}
  \bibinfo{year}{2011}\natexlab{}.
\newblock \showarticletitle{Reading digits in natural images with unsupervised
  feature learning}.
\newblock  (\bibinfo{year}{2011}).
\newblock


\bibitem[Nuriel et~al\mbox{.}(2020)]%
        {nuriel2020permuted}
\bibfield{author}{\bibinfo{person}{Oren Nuriel}, \bibinfo{person}{Sagie
  Benaim}, {and} \bibinfo{person}{Lior Wolf}.} \bibinfo{year}{2020}\natexlab{}.
\newblock \showarticletitle{Permuted adain: Enhancing the representation of
  local cues in image classifiers}.
\newblock  (\bibinfo{year}{2020}).
\newblock


\bibitem[Oh~Song et~al\mbox{.}(2016)]%
        {oh2016deep}
\bibfield{author}{\bibinfo{person}{Hyun Oh~Song}, \bibinfo{person}{Yu Xiang},
  \bibinfo{person}{Stefanie Jegelka}, {and} \bibinfo{person}{Silvio Savarese}.}
  \bibinfo{year}{2016}\natexlab{}.
\newblock \showarticletitle{Deep metric learning via lifted structured feature
  embedding}. In \bibinfo{booktitle}{\emph{Proceedings of the IEEE conference
  on computer vision and pattern recognition}}. \bibinfo{pages}{4004--4012}.
\newblock


\bibitem[Pan et~al\mbox{.}(2010)]%
        {pan2010domain}
\bibfield{author}{\bibinfo{person}{Sinno~Jialin Pan}, \bibinfo{person}{Ivor~W
  Tsang}, \bibinfo{person}{James~T Kwok}, {and} \bibinfo{person}{Qiang Yang}.}
  \bibinfo{year}{2010}\natexlab{}.
\newblock \showarticletitle{Domain adaptation via transfer component analysis}.
\newblock \bibinfo{journal}{\emph{IEEE transactions on neural networks}}
  (\bibinfo{year}{2010}), \bibinfo{pages}{199--210}.
\newblock


\bibitem[Piratla et~al\mbox{.}(2020)]%
        {piratla2020efficient}
\bibfield{author}{\bibinfo{person}{Vihari Piratla}, \bibinfo{person}{Praneeth
  Netrapalli}, {and} \bibinfo{person}{Sunita Sarawagi}.}
  \bibinfo{year}{2020}\natexlab{}.
\newblock \showarticletitle{Efficient domain generalization via common-specific
  low-rank decomposition}. In \bibinfo{booktitle}{\emph{International
  Conference on Machine Learning}}. \bibinfo{pages}{7728--7738}.
\newblock


\bibitem[Qian et~al\mbox{.}(2022)]%
        {qian2022kda}
\bibfield{author}{\bibinfo{person}{Qi Qian}, \bibinfo{person}{Hao Li}, {and}
  \bibinfo{person}{Juhua Hu}.} \bibinfo{year}{2022}\natexlab{}.
\newblock \showarticletitle{Improved Knowledge Distillation via Full Kernel
  Matrix Transfer}. In \bibinfo{booktitle}{\emph{SIAM International Conference
  on Data Mining, {SDM} 2022}}.
\newblock


\bibitem[Qian et~al\mbox{.}(2019a)]%
        {QianSSHTLJ19}
\bibfield{author}{\bibinfo{person}{Qi Qian}, \bibinfo{person}{Lei Shang},
  \bibinfo{person}{Baigui Sun}, \bibinfo{person}{Juhua Hu},
  \bibinfo{person}{Tacoma Tacoma}, \bibinfo{person}{Hao Li}, {and}
  \bibinfo{person}{Rong Jin}.} \bibinfo{year}{2019}\natexlab{a}.
\newblock \showarticletitle{SoftTriple Loss: Deep Metric Learning Without
  Triplet Sampling}. In \bibinfo{booktitle}{\emph{International Conference on
  Computer Vision}}. \bibinfo{publisher}{{IEEE}}, \bibinfo{pages}{6449--6457}.
\newblock
\urldef\tempurl%
\url{https://doi.org/10.1109/ICCV.2019.00655}
\showDOI{\tempurl}


\bibitem[Qian et~al\mbox{.}(2019b)]%
        {QianZTJSL19}
\bibfield{author}{\bibinfo{person}{Qi Qian}, \bibinfo{person}{Shenghuo Zhu},
  \bibinfo{person}{Jiasheng Tang}, \bibinfo{person}{Rong Jin},
  \bibinfo{person}{Baigui Sun}, {and} \bibinfo{person}{Hao Li}.}
  \bibinfo{year}{2019}\natexlab{b}.
\newblock \showarticletitle{Robust Optimization over Multiple Domains}. In
  \bibinfo{booktitle}{\emph{The Thirty-Third {AAAI} Conference on Artificial
  Intelligence}}. \bibinfo{publisher}{{AAAI} Press},
  \bibinfo{pages}{4739--4746}.
\newblock
\urldef\tempurl%
\url{https://doi.org/10.1609/aaai.v33i01.33014739}
\showDOI{\tempurl}


\bibitem[Qiao et~al\mbox{.}(2020)]%
        {qiao2020learning}
\bibfield{author}{\bibinfo{person}{Fengchun Qiao}, \bibinfo{person}{Long Zhao},
  {and} \bibinfo{person}{Xi Peng}.} \bibinfo{year}{2020}\natexlab{}.
\newblock \showarticletitle{Learning to learn single domain generalization}. In
  \bibinfo{booktitle}{\emph{Proceedings of the IEEE/CVF Conference on Computer
  Vision and Pattern Recognition}}. \bibinfo{pages}{12556--12565}.
\newblock


\bibitem[Saito et~al\mbox{.}(2019)]%
        {saito2019semi}
\bibfield{author}{\bibinfo{person}{Kuniaki Saito}, \bibinfo{person}{Donghyun
  Kim}, \bibinfo{person}{Stan Sclaroff}, \bibinfo{person}{Trevor Darrell},
  {and} \bibinfo{person}{Kate Saenko}.} \bibinfo{year}{2019}\natexlab{}.
\newblock \showarticletitle{Semi-supervised domain adaptation via minimax
  entropy}. In \bibinfo{booktitle}{\emph{Proceedings of the IEEE/CVF
  International Conference on Computer Vision}}. \bibinfo{pages}{8050--8058}.
\newblock


\bibitem[Shankar et~al\mbox{.}(2018)]%
        {shankar2018generalizing}
\bibfield{author}{\bibinfo{person}{Shiv Shankar}, \bibinfo{person}{Vihari
  Piratla}, \bibinfo{person}{Soumen Chakrabarti}, \bibinfo{person}{Siddhartha
  Chaudhuri}, \bibinfo{person}{Preethi Jyothi}, {and} \bibinfo{person}{Sunita
  Sarawagi}.} \bibinfo{year}{2018}\natexlab{}.
\newblock \showarticletitle{Generalizing across domains via cross-gradient
  training}.
\newblock \bibinfo{journal}{\emph{arXiv preprint arXiv:1804.10745}}
  (\bibinfo{year}{2018}).
\newblock


\bibitem[Shao et~al\mbox{.}(2019)]%
        {shao2019multi}
\bibfield{author}{\bibinfo{person}{Rui Shao}, \bibinfo{person}{Xiangyuan Lan},
  \bibinfo{person}{Jiawei Li}, {and} \bibinfo{person}{Pong~C Yuen}.}
  \bibinfo{year}{2019}\natexlab{}.
\newblock \showarticletitle{Multi-adversarial discriminative deep domain
  generalization for face presentation attack detection}. In
  \bibinfo{booktitle}{\emph{Proceedings of the IEEE/CVF Conference on Computer
  Vision and Pattern Recognition}}. \bibinfo{pages}{10023--10031}.
\newblock


\bibitem[Shi et~al\mbox{.}(2020)]%
        {shi2020informative}
\bibfield{author}{\bibinfo{person}{Baifeng Shi}, \bibinfo{person}{Dinghuai
  Zhang}, \bibinfo{person}{Qi Dai}, \bibinfo{person}{Zhanxing Zhu},
  \bibinfo{person}{Yadong Mu}, {and} \bibinfo{person}{Jingdong Wang}.}
  \bibinfo{year}{2020}\natexlab{}.
\newblock \showarticletitle{Informative dropout for robust representation
  learning: A shape-bias perspective}. In
  \bibinfo{booktitle}{\emph{International Conference on Machine Learning}}.
  \bibinfo{pages}{8828--8839}.
\newblock


\bibitem[Sun et~al\mbox{.}(2016)]%
        {sun2016return}
\bibfield{author}{\bibinfo{person}{Baochen Sun}, \bibinfo{person}{Jiashi Feng},
  {and} \bibinfo{person}{Kate Saenko}.} \bibinfo{year}{2016}\natexlab{}.
\newblock \showarticletitle{Return of frustratingly easy domain adaptation}. In
  \bibinfo{booktitle}{\emph{Proceedings of the AAAI Conference on Artificial
  Intelligence}}, Vol.~\bibinfo{volume}{30}.
\newblock


\bibitem[Sze et~al\mbox{.}(2017)]%
        {sze2017efficient}
\bibfield{author}{\bibinfo{person}{Vivienne Sze}, \bibinfo{person}{Yu-Hsin
  Chen}, \bibinfo{person}{Tien-Ju Yang}, {and} \bibinfo{person}{Joel~S Emer}.}
  \bibinfo{year}{2017}\natexlab{}.
\newblock \showarticletitle{Efficient processing of deep neural networks: A
  tutorial and survey}.
\newblock \bibinfo{journal}{\emph{Proc. IEEE}} (\bibinfo{year}{2017}),
  \bibinfo{pages}{2295--2329}.
\newblock


\bibitem[Van~der Maaten and Hinton(2008)]%
        {van2008visualizing}
\bibfield{author}{\bibinfo{person}{Laurens Van~der Maaten} {and}
  \bibinfo{person}{Geoffrey Hinton}.} \bibinfo{year}{2008}\natexlab{}.
\newblock \showarticletitle{Visualizing data using t-SNE.}
\newblock \bibinfo{journal}{\emph{Journal of machine learning research}}
  \bibinfo{volume}{9}, \bibinfo{number}{11} (\bibinfo{year}{2008}).
\newblock


\bibitem[Venkateswara et~al\mbox{.}(2017)]%
        {venkateswara2017deep}
\bibfield{author}{\bibinfo{person}{Hemanth Venkateswara}, \bibinfo{person}{Jose
  Eusebio}, \bibinfo{person}{Shayok Chakraborty}, {and}
  \bibinfo{person}{Sethuraman Panchanathan}.} \bibinfo{year}{2017}\natexlab{}.
\newblock \showarticletitle{Deep hashing network for unsupervised domain
  adaptation}. In \bibinfo{booktitle}{\emph{Proceedings of the IEEE conference
  on computer vision and pattern recognition}}. \bibinfo{pages}{5018--5027}.
\newblock


\bibitem[Wang et~al\mbox{.}(2022)]%
        {wang2022attention}
\bibfield{author}{\bibinfo{person}{Mengzhu Wang}, \bibinfo{person}{Shan An},
  \bibinfo{person}{Xiao Luo}, \bibinfo{person}{Xiong Peng},
  \bibinfo{person}{Wei Yu}, \bibinfo{person}{Junyang Chen}, {and}
  \bibinfo{person}{Zhigang Luo}.} \bibinfo{year}{2022}\natexlab{}.
\newblock \showarticletitle{Attention-based Adversarial Partial Domain
  Adaptation}. In \bibinfo{booktitle}{\emph{ICASSP 2022-2022 IEEE International
  Conference on Acoustics, Speech and Signal Processing (ICASSP)}}. IEEE,
  \bibinfo{pages}{3144--3148}.
\newblock


\bibitem[Wang et~al\mbox{.}(2021b)]%
        {wang2021interbn}
\bibfield{author}{\bibinfo{person}{Mengzhu Wang}, \bibinfo{person}{Wei Wang},
  \bibinfo{person}{Baopu Li}, \bibinfo{person}{Xiang Zhang},
  \bibinfo{person}{Long Lan}, \bibinfo{person}{Huibin Tan},
  \bibinfo{person}{Tianyi Liang}, \bibinfo{person}{Wei Yu}, {and}
  \bibinfo{person}{Zhigang Luo}.} \bibinfo{year}{2021}\natexlab{b}.
\newblock \showarticletitle{Interbn: Channel fusion for adversarial
  unsupervised domain adaptation}. In \bibinfo{booktitle}{\emph{Proceedings of
  the 29th ACM international conference on multimedia}}.
  \bibinfo{pages}{3691--3700}.
\newblock


\bibitem[Wang et~al\mbox{.}(2020)]%
        {wang2020learning}
\bibfield{author}{\bibinfo{person}{Shujun Wang}, \bibinfo{person}{Lequan Yu},
  \bibinfo{person}{Caizi Li}, \bibinfo{person}{Chi-Wing Fu}, {and}
  \bibinfo{person}{Pheng-Ann Heng}.} \bibinfo{year}{2020}\natexlab{}.
\newblock \showarticletitle{Learning from extrinsic and intrinsic supervisions
  for domain generalization}. In \bibinfo{booktitle}{\emph{European Conference
  on Computer Vision}}. \bibinfo{pages}{159--176}.
\newblock


\bibitem[Wang et~al\mbox{.}(2021a)]%
        {wang2021regularizing}
\bibfield{author}{\bibinfo{person}{Yulin Wang}, \bibinfo{person}{Gao Huang},
  \bibinfo{person}{Shiji Song}, \bibinfo{person}{Xuran Pan},
  \bibinfo{person}{Yitong Xia}, {and} \bibinfo{person}{Cheng Wu}.}
  \bibinfo{year}{2021}\natexlab{a}.
\newblock \showarticletitle{Regularizing deep networks with semantic data
  augmentation}.
\newblock \bibinfo{journal}{\emph{IEEE Transactions on Pattern Analysis and
  Machine Intelligence}} (\bibinfo{year}{2021}).
\newblock


\bibitem[Weiss et~al\mbox{.}(2016)]%
        {weiss2016survey}
\bibfield{author}{\bibinfo{person}{Karl Weiss}, \bibinfo{person}{Taghi~M
  Khoshgoftaar}, {and} \bibinfo{person}{DingDing Wang}.}
  \bibinfo{year}{2016}\natexlab{}.
\newblock \showarticletitle{A survey of transfer learning}.
\newblock \bibinfo{journal}{\emph{Journal of Big data}} (\bibinfo{year}{2016}),
  \bibinfo{pages}{1--40}.
\newblock


\bibitem[Williams and Seeger(2000)]%
        {WilliamsS00}
\bibfield{author}{\bibinfo{person}{Christopher K.~I. Williams} {and}
  \bibinfo{person}{Matthias~W. Seeger}.} \bibinfo{year}{2000}\natexlab{}.
\newblock \showarticletitle{Using the Nystr{\"{o}}m Method to Speed Up Kernel
  Machines}. In \bibinfo{booktitle}{\emph{Advances in Neural Information
  Processing Systems}}, \bibfield{editor}{\bibinfo{person}{Todd~K. Leen},
  \bibinfo{person}{Thomas~G. Dietterich}, {and} \bibinfo{person}{Volker Tresp}}
  (Eds.). \bibinfo{publisher}{{MIT} Press}, \bibinfo{pages}{682--688}.
\newblock
\urldef\tempurl%
\url{https://proceedings.neurips.cc/paper/2000/hash/19de10adbaa1b2ee13f77f679fa1483a-Abstract.html}
\showURL{%
\tempurl}


\bibitem[Xu et~al\mbox{.}(2021)]%
        {xu2021fourier}
\bibfield{author}{\bibinfo{person}{Qinwei Xu}, \bibinfo{person}{Ruipeng Zhang},
  \bibinfo{person}{Ya Zhang}, \bibinfo{person}{Yanfeng Wang}, {and}
  \bibinfo{person}{Qi Tian}.} \bibinfo{year}{2021}\natexlab{}.
\newblock \showarticletitle{A fourier-based framework for domain
  generalization}. In \bibinfo{booktitle}{\emph{Proceedings of the IEEE/CVF
  Conference on Computer Vision and Pattern Recognition}}.
  \bibinfo{pages}{14383--14392}.
\newblock


\bibitem[Xu et~al\mbox{.}(2020)]%
        {xu2020robust}
\bibfield{author}{\bibinfo{person}{Zhenlin Xu}, \bibinfo{person}{Deyi Liu},
  \bibinfo{person}{Junlin Yang}, \bibinfo{person}{Colin Raffel}, {and}
  \bibinfo{person}{Marc Niethammer}.} \bibinfo{year}{2020}\natexlab{}.
\newblock \showarticletitle{Robust and generalizable visual representation
  learning via random convolutions}.
\newblock \bibinfo{journal}{\emph{arXiv preprint arXiv:2007.13003}}
  (\bibinfo{year}{2020}).
\newblock


\bibitem[Yang et~al\mbox{.}(2021a)]%
        {yang2021adversarial}
\bibfield{author}{\bibinfo{person}{Fu-En Yang}, \bibinfo{person}{Yuan-Chia
  Cheng}, \bibinfo{person}{Zu-Yun Shiau}, {and}
  \bibinfo{person}{Yu-Chiang~Frank Wang}.} \bibinfo{year}{2021}\natexlab{a}.
\newblock \showarticletitle{Adversarial Teacher-Student Representation Learning
  for Domain Generalization}.
\newblock \bibinfo{journal}{\emph{Advances in Neural Information Processing
  Systems}}  \bibinfo{volume}{34} (\bibinfo{year}{2021}).
\newblock


\bibitem[Yang et~al\mbox{.}(2020a)]%
        {yang2020tree}
\bibfield{author}{\bibinfo{person}{Xun Yang}, \bibinfo{person}{Jianfeng Dong},
  \bibinfo{person}{Yixin Cao}, \bibinfo{person}{Xun Wang},
  \bibinfo{person}{Meng Wang}, {and} \bibinfo{person}{Tat-Seng Chua}.}
  \bibinfo{year}{2020}\natexlab{a}.
\newblock \showarticletitle{Tree-augmented cross-modal encoding for
  complex-query video retrieval}. In \bibinfo{booktitle}{\emph{Proceedings of
  the 43rd international ACM SIGIR conference on research and development in
  information retrieval}}. \bibinfo{pages}{1339--1348}.
\newblock


\bibitem[Yang et~al\mbox{.}(2020b)]%
        {yang2020learning}
\bibfield{author}{\bibinfo{person}{Xun Yang}, \bibinfo{person}{Xiaoyu Du},
  {and} \bibinfo{person}{Meng Wang}.} \bibinfo{year}{2020}\natexlab{b}.
\newblock \showarticletitle{Learning to match on graph for fashion
  compatibility modeling}. In \bibinfo{booktitle}{\emph{Proceedings of the AAAI
  Conference on Artificial Intelligence}}, Vol.~\bibinfo{volume}{34}.
  \bibinfo{pages}{287--294}.
\newblock


\bibitem[Yang et~al\mbox{.}(2021b)]%
        {yang2021deconfounded}
\bibfield{author}{\bibinfo{person}{Xun Yang}, \bibinfo{person}{Fuli Feng},
  \bibinfo{person}{Wei Ji}, \bibinfo{person}{Meng Wang}, {and}
  \bibinfo{person}{Tat-Seng Chua}.} \bibinfo{year}{2021}\natexlab{b}.
\newblock \showarticletitle{Deconfounded video moment retrieval with causal
  intervention}. In \bibinfo{booktitle}{\emph{Proceedings of the 44th
  International ACM SIGIR Conference on Research and Development in Information
  Retrieval}}. \bibinfo{pages}{1--10}.
\newblock


\bibitem[Yang et~al\mbox{.}(2020c)]%
        {yang2020weakly}
\bibfield{author}{\bibinfo{person}{Xun Yang}, \bibinfo{person}{Xueliang Liu},
  \bibinfo{person}{Meng Jian}, \bibinfo{person}{Xinjian Gao}, {and}
  \bibinfo{person}{Meng Wang}.} \bibinfo{year}{2020}\natexlab{c}.
\newblock \showarticletitle{Weakly-supervised video object grounding by
  exploring spatio-temporal contexts}. In \bibinfo{booktitle}{\emph{Proceedings
  of the 28th ACM international conference on multimedia}}.
  \bibinfo{pages}{1939--1947}.
\newblock


\bibitem[Yang et~al\mbox{.}(2018)]%
        {yang2018person}
\bibfield{author}{\bibinfo{person}{Xun Yang}, \bibinfo{person}{Peicheng Zhou},
  {and} \bibinfo{person}{Meng Wang}.} \bibinfo{year}{2018}\natexlab{}.
\newblock \showarticletitle{Person reidentification via structural deep metric
  learning}.
\newblock \bibinfo{journal}{\emph{IEEE transactions on neural networks and
  learning systems}} \bibinfo{volume}{30}, \bibinfo{number}{10}
  (\bibinfo{year}{2018}), \bibinfo{pages}{2987--2998}.
\newblock


\bibitem[Yao et~al\mbox{.}(2015)]%
        {yao2015semi}
\bibfield{author}{\bibinfo{person}{Ting Yao}, \bibinfo{person}{Yingwei Pan},
  \bibinfo{person}{Chong-Wah Ngo}, \bibinfo{person}{Houqiang Li}, {and}
  \bibinfo{person}{Tao Mei}.} \bibinfo{year}{2015}\natexlab{}.
\newblock \showarticletitle{Semi-supervised domain adaptation with subspace
  learning for visual recognition}. In \bibinfo{booktitle}{\emph{Proceedings of
  the IEEE conference on Computer Vision and Pattern Recognition}}.
  \bibinfo{pages}{2142--2150}.
\newblock


\bibitem[Zhao et~al\mbox{.}(2020)]%
        {zhao2020domain}
\bibfield{author}{\bibinfo{person}{Shanshan Zhao}, \bibinfo{person}{Mingming
  Gong}, \bibinfo{person}{Tongliang Liu}, \bibinfo{person}{Huan Fu}, {and}
  \bibinfo{person}{Dacheng Tao}.} \bibinfo{year}{2020}\natexlab{}.
\newblock \showarticletitle{Domain generalization via entropy regularization}.
\newblock \bibinfo{journal}{\emph{Advances in Neural Information Processing
  Systems}} (\bibinfo{year}{2020}), \bibinfo{pages}{16096--16107}.
\newblock


\bibitem[Zhou et~al\mbox{.}(2020a)]%
        {zhou2020deep}
\bibfield{author}{\bibinfo{person}{Kaiyang Zhou}, \bibinfo{person}{Yongxin
  Yang}, \bibinfo{person}{Timothy Hospedales}, {and} \bibinfo{person}{Tao
  Xiang}.} \bibinfo{year}{2020}\natexlab{a}.
\newblock \showarticletitle{Deep domain-adversarial image generation for domain
  generalisation}. In \bibinfo{booktitle}{\emph{Proceedings of the AAAI
  Conference on Artificial Intelligence}}. \bibinfo{pages}{13025--13032}.
\newblock


\bibitem[Zhou et~al\mbox{.}(2020b)]%
        {zhou2020learning}
\bibfield{author}{\bibinfo{person}{Kaiyang Zhou}, \bibinfo{person}{Yongxin
  Yang}, \bibinfo{person}{Timothy Hospedales}, {and} \bibinfo{person}{Tao
  Xiang}.} \bibinfo{year}{2020}\natexlab{b}.
\newblock \showarticletitle{Learning to generate novel domains for domain
  generalization}. In \bibinfo{booktitle}{\emph{European conference on computer
  vision}}. Springer, \bibinfo{pages}{561--578}.
\newblock


\bibitem[Zhou et~al\mbox{.}(2021)]%
        {zhou2021domain}
\bibfield{author}{\bibinfo{person}{Kaiyang Zhou}, \bibinfo{person}{Yongxin
  Yang}, \bibinfo{person}{Yu Qiao}, {and} \bibinfo{person}{Tao Xiang}.}
  \bibinfo{year}{2021}\natexlab{}.
\newblock \showarticletitle{Domain generalization with mixstyle}.
\newblock \bibinfo{journal}{\emph{arXiv preprint arXiv:2104.02008}}
  (\bibinfo{year}{2021}).
\newblock


\end{thebibliography}

\end{document}